# LightGBM robust optimization algorithm based on topological data analysis

Han Yang

Smart City College, Beijing Union University, 100101,2399135667@qq. com

Guangjun Qin*

Smart City College, Beijing Union University,100101, zhtguangjun@buu.edu.cn

Ziyuan Liu

Smart City College, Beijing Union University,100101, liuziyuan_cs_lzy@163.com

Yongqing Hu

Smart City College, Beijing Union University, 100101,20221081210208@buu.edu.cn

Qinglong Dai

Smart City College, Beijing Union University, 100101,xxtqinglong@buu.edu.cn

To enhance the robustness of the Light Gradient Boosting Machine (LightGBM) algorithm for image classification, a topological data analysis (TDA)-based robustness optimization algorithm for LightGBM, TDA-LightGBM, is proposed to address the interference of noise on image classification. Initially, the method partitions the feature engineering process into two streams: pixel feature stream and topological feature stream for feature extraction respectively. Subsequently, these pixel and topological features are amalgamated into a comprehensive feature vector, serving as the input for LightGBM in image classification tasks. This fusion of features not only encompasses traditional feature engineering methodologies but also harnesses topological structure information to more accurately encapsulate the intrinsic features of the image. The objective is to surmount challenges related to unstable feature extraction and diminished classification accuracy induced by data noise in conventional image processing. Experimental findings substantiate that TDA-LightGBM achieves a 3% accuracy improvement over LightGBM on the SOCOFing dataset across five classification tasks under noisy conditions. In noise-free scenarios, TDA-LightGBM exhibits a 0.5% accuracy enhancement over LightGBM on two classification tasks, achieving a remarkable accuracy of 99.8%. Furthermore, the method elevates the classification accuracy of the Ultrasound Breast Images for Breast Cancer dataset and the Masked CASIA WebFace dataset by 6% and 15%, respectively, surpassing LightGBM in the presence of noise. These empirical results underscore the efficacy of the TDA-LightGBM approach in fortifying the robustness of LightGBM by integrating topological features, thereby augmenting the performance of image classification tasks amidst data perturbations.

## 1 INTRODUCTION

Light Gradient Boosting Machine (LightGBM) is an efficient Gradient Boosting Decision Tree (GBDT) algorithmic framework[2,1], renowned for its exceptional performance in managing large-scale datasets and high-dimensional features. Unlike conventional GBDT algorithms, LightGBM integrates learning and histogram binning techniques (i.e., feature discretization)[3] to enhance model robustness, diminish memory usage, expedite training, and excel in tasks encompassing classification, regression, and ranking. Nevertheless, owing to its bias-based algorithmic approach, LightGBM exhibits heightened sensitivity to noise, potentially leading to erratic performance[4]. Thus, improving the model's ability and robustness against noise is one of the important challenges that the LightGBM algorithm needs to address.

In recent years, Topological Data Analysis(TDA)[5], a topological data analysis method based on algebraic topology theory, which can effectively capture the topological features of data under different representations, can better portray the essential structure and shape embedded in the data, and has high noise immunity, has attracted widespread attention. TDA predominantly relies on Persistent Homology (PH)[6], which accentuates global macroscopic features over local microscopic ones, encompassing phenomena like connected branches, voids, loops, and other topologically invariant features. By observing the persistence of these features across different scales, TDA unveils the intrinsic structure of the data and elucidates the significance and persistence of its topological attributes through visualization[9,8,7], This not only offers novel perspectives but also furnishes methodologies for tasks including data analysis, pattern recognition, and decision-making, thereby finding widespread applications in domains such as biology, cybersecurity, and finance.

In this paper, LightGBM is combined with TDA to propose a method called TDA-LightGBM. This method enhances feature richness by dividing feature engineering into pixel feature flow and topological feature flow, consequently enhancing the robustness of LightGBM. The pixel feature flow structures the image into feature vectors to capture the pixel information, while the topological feature flow mines the topological features of the data using the TDA technique to reveal the hidden intrinsic structure of the data. Then, the two features are aggregated and used as inputs to the LightGBM model for classification tasks to enhance model robustness.

## 2  RELATED WORK

### 2.1  LightGBM Algorithm

In the presence of noisy or fluctuating data, the LightGBM model shows sensitivity to the noisy data, which can lead to a decrease in model performance or affect its generalization ability. To enhance the robustness and stability of the model, the LightGBM model is typically utilized in conjunction with other models in practical applications. This approach maximizes the strengths of different models to construct a more robust integrated model.

Zhang Y[10]improved the stock volatility prediction system by addressing the insufficient description of the complex nonlinear features of stocks. This was achieved by inputting the feature-engineered data into LightGBM and Feed Forward Neural Network (FFNN) separately and then averaging the prediction results of the two models. Z. Cui et al.[11] decomposed and reconstructed the urban rainfall data by Singular Spectrum Analysis (SSA), and simulated the trend and fluctuation terms by using LightGBM, which successfully overcame the sensitivity to the changes of high-resolution data, and achieved real-time and accurate prediction of urban rainfall. Y. Zhao et al.[12]utilized wavelet denoising, ResNet, and LightGBM to develop a prediction model for daily trading data in financial markets. The technical indexes, processed by wavelet denoising, guided LightGBM in making predictions on feature data processed by ResNet, resulting in an enhanced accuracy rate. Hamed et al.[13]proposed a CNN-LightGBM method for cancer diagnosis. In this method, CNN was utilized for feature extraction, and LightGBM was employed for classification. This approach enhanced the model's robustness, achieving 99.6% accuracy and sensitivity in classification.

When dealing with complex real-world problems, the multi-model combination strategy of LightGBM models provides an effective approach to enhance performance and robustness. Traditional feature engineering and machine learning approaches, however, are more limited when dealing with high-dimensional, incomplete, and noise-influenced data. Introducing topological data analysis would be a powerful complement to revealing the intrinsic structure and key information of the data. By mapping the data to a topological space, it captures the potential high-level structure in the data, enhancing the model's ability to recognize complex relationships. Therefore, this paper aims to explore a methodology that combines LightGBM and TDA to construct a more comprehensive, accurate, and robust model that enhances the noise immunity of the LightGBM model.

**2.2　Topological Data Analysis**

*2.2.1　Fundamentals of Topological Data Analysis*

TDA utilizes a topological approach to map data to a topological space and extracts topological features of the data by calculating topological invariants (e.g., Betti number, persistence, etc.).These topological features help to reveal the intrinsic structure and key information of the data, and TDA quantifies and visualizes these features in the form of Betti curves[14], Persistence Diagram (PD)[15], etc [16]. Among them, the Betti number reflects the number of topological features in various dimensions, while the Persistence Diagram reflects the persistence of these topological features.

The Betti number specifically refers to the number of holes in a topological space in various dimensions. It is utilized to quantify the topological characteristics of the space and is a significant topological invariant. The 0-dimensional Betti number denotes the number of connected components, the 1-dimensional Betti number denotes the number of rings, the 2-dimensional Betti number denotes the number of voids in the space, and so on. Furthermore, Betti curves can illustrate the trend of Betti numbers at various parameter values.

Persistence Diagrams[18,17] are two-dimensional graphs that reflect topological information at different scales by visualizing the generation and disappearance times of each topological feature. Figure 1 (b) is the persistence diagram obtained from the topological data analysis in Figure 1 (a). As shown in Figure. 1(b), "H0" stands for 0-dimensional topological features, usually denoting connected components or isolated points. H1 stands for a 1-dimensional topological feature, typically representing shapes like rings and holes. The X-coordinate represents the birth value (the time when the topological features started to be generated), and the Y-coordinate represents the death value (the time when the topological feature ended). All points lie on the diagonal because topological features always disappear after birth. The persistence of a feature is represented by the distance from the point to the diagonal. Longer durations indicate more stable topological features, while short durations are considered noisy or unstable features. Persistence maps serve as a powerful tool that not only provides a visual means for the topological analysis of data but also offers a way to gain a deeper understanding of the intrinsic structure of the data.

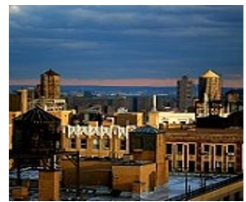
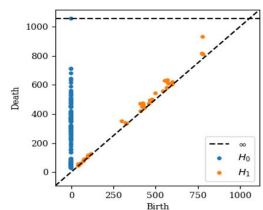

（a）image　　　（b）Persistence Diagram

Figure 1 Images are analyzed for topological data to obtain persistence Diagram

In tasks such as image classification, feature stability is crucial for model performance. Real-world images are often affected by various noises and disturbances, which can degrade the model's classification performance. Compared to "black-box" methods (e.g., deep neural networks)[19], TDA demonstrates excellent noise immunity when handling high-dimensional, incomplete, and noise-disturbed images[21,20,15,5], TDA is capable of extracting stable topological features that exhibit good stability and robustness to data perturbations.

*2.2.2 Topological Data Analysis with machine learning and deep learning*

TDA is effective in guiding machine learning models for image data analysis, providing insights against complex structures and changing patterns in image data. Muszynski et al.[21] combining TDA and Support Vector Machine (SVM) to automatically identified topological features of atmospheric rivers ( Atmospheric Rivers, ARs) topological features to address the uncertainty of weather patterns and achieve high-precision recognition of atmospheric river patterns. Schiff Y et al.[16] utilized a machine learning approach combined with geometric features extracted by TDA to classify images from mouse blood and lymphatic vessel network imaging systems. They employed a non-chiral Poisson process (NHPP) model to simulate vessel strength, calculated distance matrices using Sliced Wasserstein, and visualized similarities between regions using Multidimensional Scaling (MDS) to assess differences in lymphatic vessel structure. Thomas et al.[22] used ridge regression in conjunction with topological features extracted by radionics and TDA for malignancy prediction. This approach significantly improved the model performance and successfully addressed the issue of the risk stratification system (RSS) being unable to accurately differentiate follicular carcinoma from adenoma.

TDA and Convolutional Neural Network (CNN) models[23] were combined to achieve a comprehensive understanding of image data by integrating the feature representations learned from the image by CNN and the topological structure information extracted by TDA. Hajij M et al.[24] proposed TDA-Net, which comprises a deep branch and a topological branch. It fuses the features learned by the deep neural network with the topological features extracted by TDA. This fusion approach achieved the best classification results by combining the outputs of the two branches for the classification task. Elyasi et al.[25] utilized the Xception network to extract features and then integrated them with features extracted by a deep learning model. The combined features were processed by the TDA Mapper to extract additional topological features, leading to the successful classification of skin cancer images. Kun Wang et al.[26] proposed the TCAG model, which utilizes a convolutional neural network to extract spatiotemporal features of the data, applies the TDA method to extract the topological features of the time series and then inputs the combination of these two features into the Gated Recurrent Unit (GRU) neural network to improve the accuracy of fault prediction.

Overall, TDA introduces novel methods and techniques in the realm of machine learning and deep learning. It can be integrated with machine learning models or deep convolutional neural networks, yielding favorable outcomes. TDA broadens the capabilities of models to process intricate data and enhance their resilience. This provides an effective method for analyzing data to solve numerous practical problems, particularly in fields like medicine, biology, and meteorology.

## 3 LIGHTGBM ROBUST OPTIMIZATION ALGORITHM BASED ON TOPOLOGICAL DATA ANALYSIS

### 3.1 Algorithmic Architecture

To enhance the robustness of LightGBM, this paper proposes a TDA-LightGBM method. As shown in Figure 2, the process of feature engineering is divided into pixel feature flow and topological feature flow. The pixel feature flow is responsible for extracting the values of pixels from the image data and converting them into feature vector representations. The topological feature flow utilizes TDA to extract topological structural features of the image. This helps capture the hidden structural relationships in the data and provides deeper insights into the model. The two features are aggregated to form a combined feature vector, which is used as an input to the LightGBM model for classification. This enhances the model's classification performance and robustness.

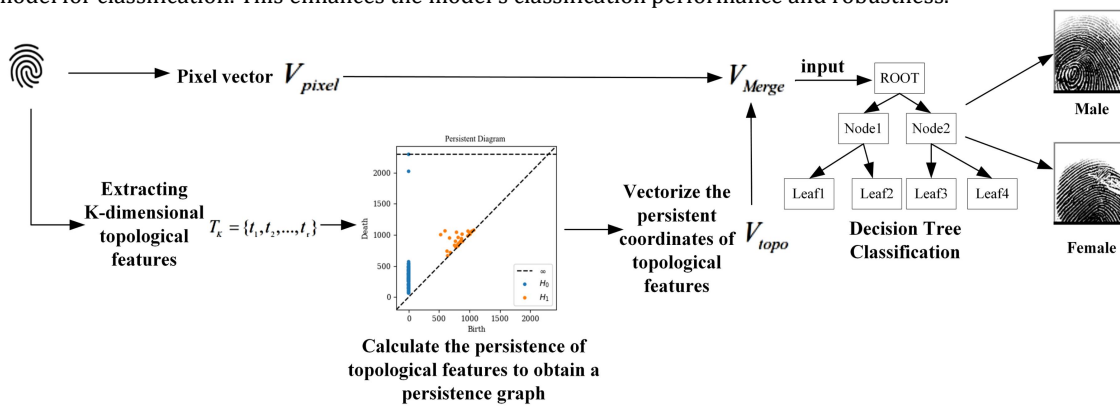

Figure 2 LightGBM Robust Optimization Algorithm Based on Topological Data Analysis (An Example of Gender Binary Classification with SOCOFing Fingerprint Dataset)

The main operations of the model in Figure 2 are as follows:

(1) Topological feature extraction: The basic principle of topological data analysis is to construct simplexes on a point cloud[27] and record the birth and death times of the simplexes[28]. One-dimensional simplexes are line segments, two-dimensional simplexes are triangles, three-dimensional simplexes are tetrahedra, and so on. They are used to describe the fundamental structure of the topological space.

Let the point cloud data of the image $D$ be the set $D = \{d_1, d_2, ..., d_n\}$, where $d_i$ denotes the data point. Carving topological relationships between data points by constructing a collection $S = \{\partial_1, \partial_2, ..., \partial_m\}$ of simplexes on the point cloud, where $\partial_K$ represents the kth dimensional simplexes. A combination of simplexes is called a simplex complex, and in this paper we use the Vietoris-Rips complex[29]. The formation process of the complex is shown in Figure 3, which performs persistence analysis to capture the topological features of the data by setting a distance threshold $\varepsilon$ between the data points.

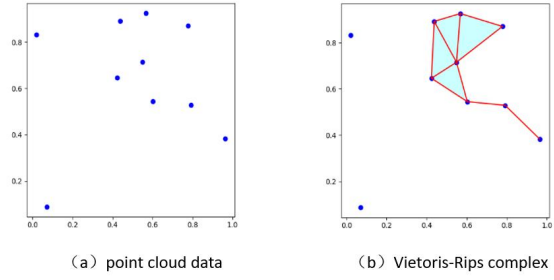

Figure 3 Formation process of the Vietoris-Rips complex shape

Let the set of obtained Kth dimensional topological features be $T_K = \{t_1, t_2, ..., t_r\}$, where $t_i$ denotes the i-th topological feature; whose persistence can be visualized by a persistence graph, denoted as $PD_K^e = \{(x_i, y_i) | i = 1, 2, ..., r\}$, where $(x_i, y_i)$ denotes the persistence of the i-th topological feature, $x_i$ denotes the birth time of the feature, and $y_i$ denotes the death time of the feature. Vectorise the persistent coordinates of these topological features to represent them in a form that is standardised and suitable for processing by machine learning algorithms, yielding a feature vector $V_{topo} = [v_1, v_2, ..., v_N]$, which $v_i$ denotes the persistence coordinate of the i-th feature.

(2) Pixel feature extraction: let the original image array be $I$, $I \in R^{H \times W \times C}$, Where $H$ is the image height, $W$ is the image width, and $C$ is the number of channels. Extract pixel values from the image data and vectorize the 3D image array $I$ into pixel vectors $V_{pixel} = [v_1, v_2, ..., v_{H \times W \times C}]$, its length is $I = H \times W \times C$.

(3) Feature fusion: Machine learning models require inputs of the same length. The length of topological features may vary from image to image, resulting in the topological feature vectors' lengths varying from image to image. In contrast, the length of the pixel vectors is fixed based on the image size after vectorization operations. To ensure that the feature vectors are of the same length and to meet the requirement of limiting the data dimensionality to a specific range for fitting the model inputs, a "truncation" strategy is employed to align the dimensional aspects of the topological features.

By conducting a persistence analysis on the topological features, the first few important topological features are selected for truncation to reduce dimensionality and preserve the features that have a more pronounced impact on the data structure. The remaining features will be truncated and discarded. By truncating the topological features, the length of these vectors can be adjusted to the same threshold to ensure data consistency. In the experiments, choosing the truncation location is a critical step. The optimal truncation location needs to be determined several times through experimental tuning to ensure that the retained features can capture the key information of the data in the most effective way.

Finally, the topological features are combined with the pixel features to form a combined feature vector denoted as $V_{Merge} = [V_{pixel}, V_{topo}]$. This fusion of pixel values and structural features offers a more comprehensive description of the information in the image, delivering richer and more informative inputs to the machine learning model. This enhancement improves model performance and prediction accuracy.

（4）Image classification: The process of inputting the aggregated feature vector $V_{Merge}$ into the LightGBM model to perform the image classification task is denoted as: $Classification = LightGBM(V_{Merge})$.

Traditional machine learning methods typically focus solely on the low-order features of the data, overlooking higher-order features like correlations and topology between data points. The TDA-LightGBM method not only considers the pixel features of the image but also captures the connectivity and topology between data points. The fusion of these two features can better reflect the multi-dimensional features of the data. The topological invariance of high-order features makes it possible to enhance the stability of the features and improve the model's performance by increasing robustness to data noise. The introduction of topological features is crucial as it enhances the interpretability of the features. Most importantly, the introduction of topological features enhances the interpretability of the features, contributing to a clearer understanding of the relationships between the features and the predictions of the model.

### 3.2 Core Algorithm

The TDA LightGBM method described in Algorithm 1 of Table 1. There is a color image dataset $D$ containing $n$ samples. Before processing the images, each color image in the dataset $D$ is first converted to a grayscale image $G$ by a function $gray$. In the algorithm, Vietoris-Rips complex shapes are constructed based on the distance parameter $\varepsilon$ between data points. Then topological features in the image data are extracted and analyzed through persistent cohomology, The importance and persistence of the features are identified by visualizing the topological features as persistence diagrams. Let $f$ denote the function that extracts topological features on the grayscale graph $G$. The generation of topological features in different dimensions is an iterative process, In each iteration, the topological features of the next dimension are generated. This process is denoted as $T_0 = f(G), T_1 = f(T_0), ..., T_K = f(T_{K-1})$; The set of k-th dimensional topological features $T_K = \{t_1, t_2, ..., t_r\}$ is obtained, where $t_i$ denotes the i-th k-dimensional topological feature. Under the distance parameter $\varepsilon$, by analyzing the persistence of topological features, the resulting k-dimensional persistence map is represented as $PD_K^\varepsilon$; Let the vectorization operation be denoted as a function of $flat$.

Table 1 Algorithm1 Description

Algorithm 1 LightGBM robust optimization algorithm based on topological data analysis

input：ɑ：Topological eigenvector truncation parameters；β：Pixel vector truncation parameters
output：$Accuracy$、$Precision$、$Recall$、$F_1$

1、 $G \leftarrow gray(D)$    //Convert to grayscale
2、 $T_K \leftarrow f(T_{K-1}), K \geq 1; T_0 \leftarrow f(G), K = 0$    //Extracting topological features
3、 $PD_k^\varepsilon \leftarrow \{(x_i, y_i) | i = 1, 2, ..., r\}$    //Obtain the persistence graph:where $(x_i, y_i)$ denotes the birth time and death time of the i-th topological feature
4、 $V_{topo} \leftarrow flat\{(x_i, y_i) | i = 1, 2, ..., r\}$    //Vectorising persistence coordinates
5、 $V_{pixel} \leftarrow flat(I)$    //Vectorise the image array $I$
6、 for each $ɑ = 0, ɑ \leq V_{topo}; β = 0, β \leq V_{pixel}$
7、     $V_{Merge} \leftarrow [ɑ \cdot V_{topo}, β \cdot V_{pixel}]$    //dimensional alignment

    8、 end
    9、 $LightGBM(V_{Merge})$    //Training Models

ALGORITHM ANALYSIS: The algorithm contains two key mapping functions for vectorizing the persistence coordinates of image arrays and topological features, respectively. These two mapping functions are denoted as: $V_{pixel} \leftarrow flat(I)$、$V_{topo} \leftarrow flat\{(x_i, y_i) | i=1,2,...,r\}$；included among these，$V_{pixel}$ denotes a vector of pixels to vectorize the image array，$V_{topo}$ denotes the topological feature vector obtained after vectorization.

To aggregate the topological features and pixel features to obtain the combined feature vector $V_{Merge}$, this process is achieved through a merge function denoted as: $V_{Merge} \leftarrow [\alpha \bullet V_{topo}, \beta \bullet V_{pixel}]$. Parameters α and β are introduced in this algorithm, representing the truncation parameters for adjusting the dimensionality of topological features and pixel features, respectively. Eventually, the adjusted combined feature vectors are fed into the LightGBM model for training, and the prediction results are obtained.

The algorithm describes the execution process of the TDA-LightGBM algorithm. By extracting topological features can gain a better understanding of the shape, structure, and topological properties of the image. In addition to the topological features, the fundamental pixel features of the image are also considered. To efficiently combine the topological feature vectors and pixel vectors efficiently, feature aggregation was performed using a truncation method. The integrated feature representation provides a more comprehensive and detailed data representation for the model. Ultimately, the aggregated feature vectors are inputted into the LightGBM model for training in image classification tasks to enhance accuracy and robustness.

## 4 EXPERIMENTAL ANALYSIS

### 4.1 Datasets and Evaluation Indicators

The experiments in this paper utilize an enhanced version of the SOCOFing dataset[30],Three levels of synthetic modifications, such as elimination, center rotation, and z-cut operations, were conducted using the STRANGE toolbox to create image datasets at three varying difficulty levels: easy, medium, and hard. These versions differ in image quality, and as the difficulty level increases, the image clarity gradually decreases. These modifications involve artificially processing the image data to simulate various image qualities and disturbances that may occur in the real world. This process can be seen as adding artificial noise to the images.

In the experiments, based on the needs of the experiments and design considerations, the $V_{topo}$ were adjusted according to the $V_{pixel}$ with the best experimental accuracy on the LightGBM model first to determine the robustness of the TDA-LightGBM approach. The step size of the image pixel vectors is set as 1000. The values between $1/5$ and $1$ of the total number of topological feature vector values are taken as the dimensions of the alignment, respectively, and are taken in steps of $1/5$. In the experiments, the optimal truncation position was determined by multiple tunings. To evaluate the performance of the modeling process, four possible types of metrics were used: True Positive（TP), False Positive（FP), True Negative（TN）,False Negative(FN). Accuracy, Precision, Recall, and F1 score were used as evaluation indicators.

### 4.2 Analysis of Results

#### 4.2.1 Comparison Results Without Added Gaussian Noise

Evaluating the performance of each difficulty version of the SOCOing dataset, it can be observed that in the Medium and Hard versions of the classification task, the accuracy of the TDA-LightGBM model slightly decreases compared to the Easy version. The magnitude of this decrease is not significant. In the Hard version of the five-classification task, the decrease in accuracy is more pronounced, but the performance is still maintained at a high standard. This demonstrates that the TDA-LightGBM model can maintain its stability and anti-interference ability even when the image quality degrades. Figure 4 illustrates the classification performance of TDA-LightGBM on the three difficulty levels of the SOCOFing dataset, with accuracy as the primary evaluation indicator.

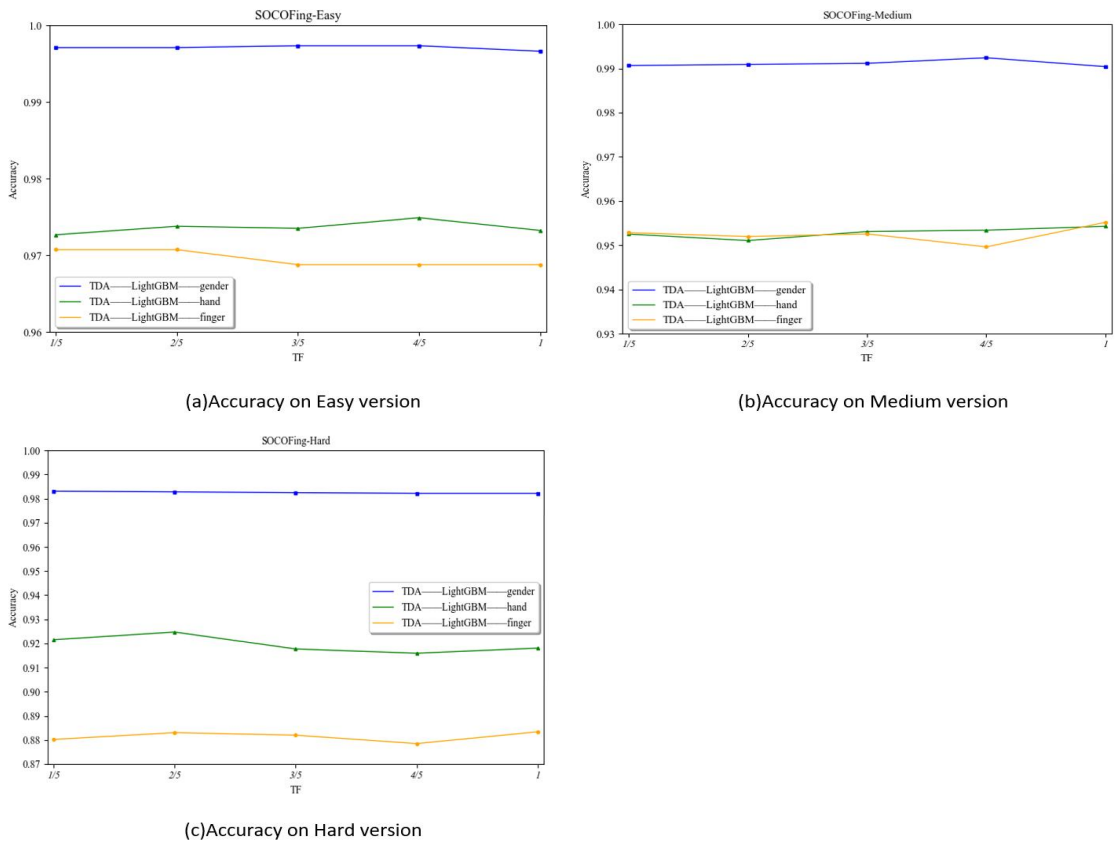

(a) Accuracy on Easy version    (b) Accuracy on Medium version

(c) Accuracy on Hard version

Figure 4  Classification accuracy of TDA-LightGBM on SOCOFing dataset

Taking the Easy version as an example, Figure 5 shows the comparison results between LightGBM and TDA-LightGBM on the classification tasks of gender binary classification, left and right-hand binary classification, and finger-five classification. The performance improvement of the TDA-LightGBM method for LightGBM is presented using accuracy as an evaluation indicator.

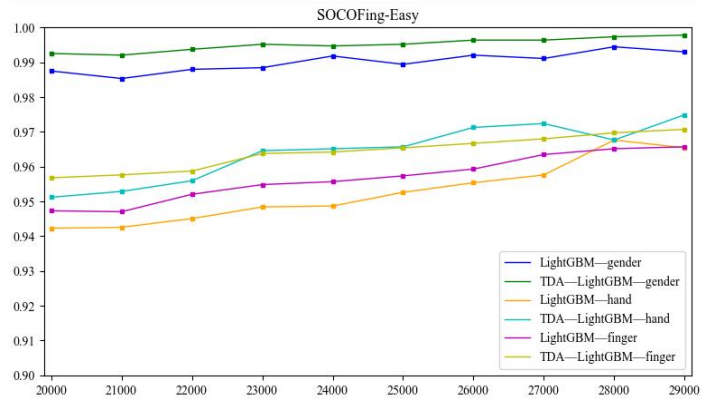

Figure 5 Experimental results of classification accuracy comparison between LightGBM and TDA-LightGBM on the Easy version

The experimental results demonstrate that while the original LightGBM model exhibited high classification accuracy, the TDA-LightGBM model, enhanced by integrating the TDA technique, achieved a higher accuracy rate than the LightGBM model in the image classification task for processing the SOCOFing dataset. Specifically, the TDA-LightGBM model's accuracy is enhanced by 0.2%, 0.7%, and 0.5% compared to the base LightGBM model in the gender binary classification, left-right hand binary classification, and finger five classification tasks, respectively.

Taking the Easy version as an example, Table 2 presents the performance of several typical machine learning algorithms used in gender binary classification tasks, including Support Vector Machine (SVM), K-Nearest Neighbor (KNN), Naive Bayes model, LightGBM, and the TDA LightGBM method proposed in this paper, evaluated by accuracy.

Table 2 Comparison of Machine Learning Algorithms

| methods | Accuracy | Precision | Recall | F1 |
|---|---|---|---|---|
| SVM | 0.966 | 0.982 | 0.968 | 0.975 |
| KNN | 0.901 | 0.948 | 0.904 | 0.925 |
| Naive Bayes | 0.382 | 0.833 | 0.111 | 0.196 |
| LightGBM | 0.994 | 0.994 | 0.998 | 0.996 |
| **TDA-LightGBM** | **0.997** | **0.996** | **1.00** | **0.998** |

Observing the comparison results, it is evident that the TDA LightGBM method exhibits significantly higher accuracy compared to other machine learning methods, showcasing superior performance in this task. This result validates the effectiveness of combining the topological features extracted by TDA technology with traditional pixel features. It emphasizes the potential benefits of incorporating image structure information revealed by TDA in image classification tasks to enhance classification performance.

*4.2.2 Adding Gaussian Noise Comparison Results*

Considering the presence of fragmented information, misinformation, and noise in image data, it can adversely affect the results of image classification.In practical applications, images are usually captured in complex real-world environments and may be affected by various influences from the surrounding environment, which increases the difficulty of processing image data. Therefore, it is important to study the effect of noise on image features and further analyze its impact on the classification results to enhance the accuracy and robustness of image classification[31].

To evaluate the robustness of the TDA-LightGBM model in a noisy environment, the most classical and practical Gaussian noise model is introduced to the image in this experimentt ($Mean = 0$, $Sigma = 0.1$) to simulate real-world noise environments, and the best values of $V_{pixel}$ and $V_{topo}$ are selected for comparison experiments, respectively, aiming to evaluate the performance of the TDA-LightGBM model in noisy environments to assess its robustness in real application scenarios. Table 3 shows the results of the comparison experiments between LightGBM and TDA-LightGBM in image classification.

Table 3 Experimental results comparing LightGBM and TDA-LightGBM classification

| datasets | Experimental Methods | Accuracy | Precision | Recall | F1 |
|---|---|---|---|---|---|
| Easy | LightGBM -gender | 0.86 | 0.84 | 0.99 | 0.90 |
| | TDA-LightGBM-gender | 0.88 | 0.85 | 1.00 | 0.92 |
| | LightGBM-hand | 0.87 | 0.88 | 0.85 | 0.87 |
| | TDA-LightGBM-hand | 0.90 | 0.91 | 0.88 | 0.89 |
| | LightGBM-finger | 0.83 | 0.83 | 0.83 | 0.83 |
| | TDA-LightGBM-finger | 0.86 | 0.86 | 0.86 | 0.86 |
| Medium | LightGBM-gender | 0.82 | 0.80 | 0.99 | 0.88 |
| | TDA-LightGBM- gender | 0.81 | 0.78 | 0.99 | 0.88 |
| | LightGBM-hand | 0.85 | 0.84 | 0.88 | 0.86 |
| | TDA-LightGBM-Hand | 0.87 | 0.88 | 0.84 | 0.86 |
| | LightGBM-finger | 0.75 | 0.75 | 0.75 | 0.75 |
| | TDA-LightGBM-finger | 0.79 | 0.79 | 0.79 | 0.79 |
| Hard | LightGBM-gender | 0.81 | 0.80 | 0.99 | 0.88 |
| | TDA-LightGBM-gender | 0.80 | 0.78 | 0.99 | 0.87 |
| | LightGBM-hand | 0.84 | 0.82 | 0.88 | 0.85 |
| | TDA-LightGBM-Hand | 0.84 | 0.85 | 0.81 | 0.83 |
| | LightGBM-finger | 0.69 | 0.69 | 0.69 | 0.69 |
| | TDA-LightGBM-finger | 0.70 | 0.70 | 0.70 | 0.70 |

The experimental results show that in most cases, the TDA-LightGBM model demonstrates superior performance in image classification tasks compared to the LightGBM model. The TDA-LightGBM model demonstrates its unique robustness against image noise by incorporating topological information.

Overall, the TDA-LightGBM method can process image data more robustly in the presence of noise. It effectively utilizes topological information to resist noise interference, thereby enhancing classification performance. This feature enables the TDA-LightGBM method to maintain good performance in complex image environments and provides a reliable solution for practical applications.

*4.2.3 Robustness Validation of the TDA-LightGBM Model*

To further validate the robustness of the TDA-LightGBM model, three datasets, CIFAR-10, Maskedd CASIA WebFace, and Ultrasound Breast Images for Breast Cancer, were used for the experiments,in storage space, an approximate number of 494,414 Masked CASIA WebFace datasets were extracted. The results of the comparison experiments on the three datasets with the addition of Gaussian noise ($M=0$, $S=0.1$) on the TDA-LightGBM and LightGBM models are presented in Table 4.

Table 4 Experimental results on multiple datasets

| datasets | Experimental Methods | Accuracy | Precision | Recall | F1 |
|---|---|---|---|---|---|
| CIFAR-10 | LightGBM | 0.49 | 0.48 | 0.49 | 0.48 |
|  | TDA-LightGBM | **0.49** | **0.45** | **0.49** | **0.44** |
| Maskedd CASIA WebFace | LightGBM | 0.16 | 0.21 | 0.16 | 0.04 |
|  | TDA-LightGBM | **0.31** | **0.33** | **0.31** | **0.31** |
| Ultrasound Breast Images for Breast Cancer | LightGBM | 0.72 | 0.76 | 0.73 | 0.74 |
|  | TDA-LightGBM | **0.78** | **0.80** | **0.82** | **0.81** |

Based on the results of the above multi-dataset experiments, it is observed that overall, TDA-LightGBM demonstrates better accuracy compared to LightGBM after the introduction of noise. However, there are significant differences in data distribution among the various datasets, leading to varying degrees of improvement of TDA-LightGBM across different datasets. On the CIFAR-10 dataset, the two models show relatively flat performance levels; however, on the Maskedd CASIA WebFace dataset and Ultrasound Breast Images for Breast Cancer dataset, the improvement of TDA-LightGBM is significant. This difference may stem from the increased difficulty in extracting topological features with the addition of noise in the CIFAR-10 dataset. The presence of noise may lead to changes in the topology, blurring some of the topological features and making it challenging to distinguish the ten categories of the images. In contrast, when considering the Maskedd CASIA WebFace dataset and Ultrasound Breast Images for Breast Cancer dataset, the addition of noise might alter the morphology of certain areas within the images. However, in comparison to the entire image, the morphology of specific regions such as the masks and breasts in the dataset changes relatively small. Consequently, the topological features remain effective in capturing the structural information of these regions accurately. Therefore, the performance improvement of the TDA-LightGBM model on these datasets is more significant. It can effectively cope with the challenges posed by noise, thus enhancing the accuracy of image classification.

By integrating the topological features of the data and the learning capability of LightGBM enables TDA-LightGBM to achieve more accurate and stable classification results in noisy environments, thereby enhancing the reliability and robustness of the model in practical applications.

## 5 CONCLUDING REMARKS

In this study, the fusion strategy of TDA and machine learning algorithms is thoroughly explored. An integrated TDA-LightGBM model is successfully constructed, which conducts feature extraction by dividing the feature engineering step into pixel feature flow and topological feature flow, respectively. The feature vectors and topological features are then combined into an integrated feature vector as the input to LightGBM. This integration enhances the accuracy of image classification. The robustness of the TDA-LightGBM model is verified through a series of experiments conducted in a noisy environment. The results show that the TDA-LightGBM method proposed in this paper performs more robustly in the image classification task in the presence of Gaussian noise influence, compared to the classification method using only LightGBM.

However, there is also room for improvement in the method proposed in this paper. When performing topological feature extraction, methods such as Principal Component Analysis (PCA) can be considered to identify the features that significantly influence the classification results. This approach enables more effective feature extraction and dimensionality reduction, leading to further optimization of the classification model's performance. Secondly, in the classification stage, the method can be enhanced by iteratively adjusting the number of leaf nodes in the decision tree to identify features that are better suited for integrating both pixel and topological features, and to establish the optimal decision strategy. Finally, inspired by conducting experiments on the fusion of TDA with LightGBM, comparative experiments on combining TDA with other machine learning and deep learning methods will be conducted in future work. The aim is to design integrated models that can address the limitations of both topological data analysis and machine learning models.

Currently, some experiments on the synergistic application of TDA with convolutional neural networks have been completed and have achieved good results. This will help to understand the image features in complex datasets more comprehensively and accurately, leading to improved performance of the models in executing the corresponding tasks. The advancement of this research direction is expected to bring more profound insights and innovations in the field of machine learning and data analysis.

## ACKNOWLEDGMENTS

We sincerely appreciate the generous funding from the Competitive Research Program (funding unit: Beijing United University Research Project Funding, project number ZKZD202301). Without their financial assistance, this study would be impossible.Thank you to all participants who voluntarily invested their time and insights into this study. Their cooperation and dedication are crucial to the success of our research.Finally, we are very grateful to our family and friends for their unwavering support and encouragement in this endeavor. Their encouragement kept us motivated during challenging times.We appreciate the collective efforts of everyone involved in this study. Without them, this work could not have been completed.